\documentclass{article}

\PassOptionsToPackage{numbers, compress}{natbib}
\usepackage[final]{nips_2017}

\usepackage[utf8]{inputenc} 
\usepackage[T1]{fontenc}    
\usepackage{hyperref}       
\usepackage{url}            
\usepackage{booktabs}       
\usepackage{amsfonts}       
\usepackage{nicefrac}       
\usepackage{microtype}      

\usepackage{amsmath,amssymb}
\usepackage{graphicx,soul,subcaption}
\usepackage{algorithm}
\usepackage{algpseudocode}

\usepackage{color}

\newcommand{\red}[1]{\textcolor{red}{#1}}
\newcommand{\fsize}{0.12}
\title{Feature Control as Intrinsic Motivation for Hierarchical Reinforcement Learning}

\author{
  Nat Dilokthanakul \\
  Imperial College London\\
  \texttt{n.dilokthanakul14@imperial.ac.uk} \\
  \And
   Christos Kaplanis \\
   Imperial College London \\
  \texttt{christos.kaplanis14@imperial.ac.uk} \\
  \AND
   Nick Pawlowski \\
   Imperial College London \\
  \texttt{n.pawlowski16@imperial.ac.uk} \\
   \And
   Murray Shanahan\\
   Imperial College London \& DeepMind\\
   \texttt{m.shanahan@imperial.ac.uk} \\
}

\begin{document}

\maketitle

\begin{abstract}

The problem of sparse rewards is one of the hardest challenges in contemporary reinforcement learning. Hierarchical reinforcement learning (HRL) tackles this problem by using a set of temporally-extended actions, or options, each of which has its own subgoal. These subgoals are normally handcrafted for specific tasks. Here, though, we introduce a generic class of subgoals with broad applicability in the visual domain. Underlying our approach (in common with work using ``auxiliary tasks'') is the hypothesis that the ability to control aspects of the environment is an inherently useful skill to have. We incorporate such subgoals in an end-to-end hierarchical reinforcement learning system and test two variants of our algorithm on a number of games from the Atari suite. We highlight the advantage of our approach in one of the hardest games -- Montezuma's revenge -- for which the ability to handle sparse rewards is key. Our agent learns several times faster than the current state-of-the-art HRL agent in this game, reaching a similar level of performance. \red{UPDATE 22/11/17: We found that a standard A3C agent with a
simple shaped reward, i.e. extrinsic reward + feature control intrinsic reward,
has comparable performance to our agent in Montezuma Revenge. In light of the new
experiments performed, the advantage of our HRL approach can be attributed more
to its ability to learn useful features from intrinsic rewards rather than its ability to
explore and reuse abstracted skills with hierarchical components. This has led
us to a new conclusion about the result.}

\end{abstract}

\section{Introduction}

Reinforcement learning methods~\cite{Sutton:1998, mnih2015human} often struggle in environments where the rewards are sparsely encountered, and when their acquisition requires the coordination of temporally extended sequences of actions. In these types of environments, archetypally exemplified by the Atari game Montezuma’s revenge, the dearth of feedback the agent receives from the environment makes it very difficult to learn long sequences of actions, particularly when the timescale of the exploration strategy is short.
 
Hierarchical reinforcement learning (HRL)~\cite{barto2003recent} is an approach that aims to deal with the reward sparsity problem by equipping the agent with temporally extended macro-actions, also known as options~\cite{sutton1999between} or skills~\cite{konidaris2009skill}, which abstract over sequences of primitive actions. If useful options are established, then long sequences of primitive actions can be expressed by much shorter sequences of options, which are easier to learn as the agent can now employ temporally extended exploration in the option space. However, learning useful options is a difficult task in itself; one possibility is to incorporate prior knowledge about the task into their construction~\cite{barto2003recent,parr1998reinforcement,dietterich2000hierarchical} but this can limit the generalisability of the algorithm to other tasks.
 
In this paper, we constrain this prior knowledge to the hypothesis that the ability to control features of its environment is an inherently useful skill for an agent to have for succeeding at a wide variety of tasks. By applying this concept in a deep HRL setting, we design an agent that is intrinsically motivated to control aspects of its environment via a set of options. 
 
The architecture of our agent is inspired by feudal reinforcement learning~\cite{dayan1992feudal,vezhnevets2017feudal} and the hierarchical deep reinforcement learning framework~\cite{kulkarni2016hierarchical}, whereby a meta-controller provides embedded subgoals to a sub-controller that interacts directly with the environment (see Figure~\ref{fig:HRL}). In our agent, the meta-controller, which learns to maximise extrinsic reward from the environment, tells the sub-controller what feature of the environment it should control; the sub-controller receives intrinsic rewards for successfully changing the chosen feature as well as extrinsic rewards from the environment. We show that, when guided by this form of intrinsic motivation, the agent learns to perform better in tasks featuring sparse rewards.

Our main contribution is in the design of discrete sets of subgoals available for the meta-controller to choose from and their corresponding intrinsic reward. By taking an existing idea of feature control~\cite{jaderberg2016reinforcement, bengio2017independently} and incorporating it into the subgoal design, we introduce a hierarchical agent with generically useful learnable options which we empirically evaluate in the Atari domain.

\subsection{Related work}

The idea of embodying an agent with a form of intrinsic motivation, which in our case is the desire to be able to control aspects of its environment, is one that has been explored in several other works. Klyubin et al. introduced \emph{empowerment} as an information theoretic measure of degrees of freedom that an agent has over an environment ~\cite{klyubin2005empowerment}. The concept of empowerment has recently gained interest in the context of intrinsically motivated reinforcement learning~\cite{mohamed2015variational, gregor2016variational}. In the other lines of work, intrinsic motivation is defined in the form of \emph{curiosity}, which can be measured with model-learning progress~\cite{schmidhuber1991curious, houthooft2016vime} or information gain~\cite{bellemare2016unifying}.     

Jaderberg et al. introduced the idea of \emph{off-policy} training with auxiliary control tasks, such as  pixel control or feature control, which can significantly speed up learning of the main task~\cite{jaderberg2016reinforcement}. The rationale is that learning the auxiliary tasks gives the agent features that are useful for manipulating the environment. Drawing on this idea, we apply the idea of pixel and feature control to the HRL framework so that options are constructed with the explicit motive of altering given features or patches of pixels. By doing so, our agent is equipped with temporally-extended options, which can be used \emph{on-policy} to explore the environment in a temporally-extended manner, thereby helping to address the problem of sparse reward.

Our architecture also takes inspiration from recent work by Kulkarni et al.~\cite{kulkarni2016hierarchical} and Vezhnevets et al.~\cite{vezhnevets2017feudal}. These works outline hierarchical architectures that comprise a subgoal-selecting meta-controller and a sub-controller that tries to achieve the subgoal. The main feature that sets our model apart is the design of the subgoals. Kulkarni et al. pre-define a set of discrete subgoals specific to the tasks at hand, while Vezhnevets et al. construct subgoals as a large continuous set of embedded states. We construct two discrete sets of subgoals, which are discussed in more detail in Section~\ref{sec:measure_control}. One of them is fixed but is designed to be generically applicable in visual domains; the other can be automatically learned such that the subgoals are useful for solving the task at hand.

There are a large number of works on subgoal discovery~\cite{csimcsek2004using, menache2002q, mcgovern2001automatic}, most of which are based on finding \emph{bottleneck} states. Since finding bottleneck states requires global statistics of the environment, finding them can be difficult and hard to scale. Our work is in line with contemporary HRL, e.g. Option-Critic~\cite{bacon2016option}, which has moved towards end-to-end training where options and subgoals can automatically emerge from the optimisation of the system, with carefully designed architectures and objective functions.

\section{The model}

\begin{figure}[t]
  \centering
  \begin{subfigure}{0.32\textwidth}
	\centering
	\includegraphics[trim={0cm 0cm 0cm 0cm},clip,width = \textwidth]{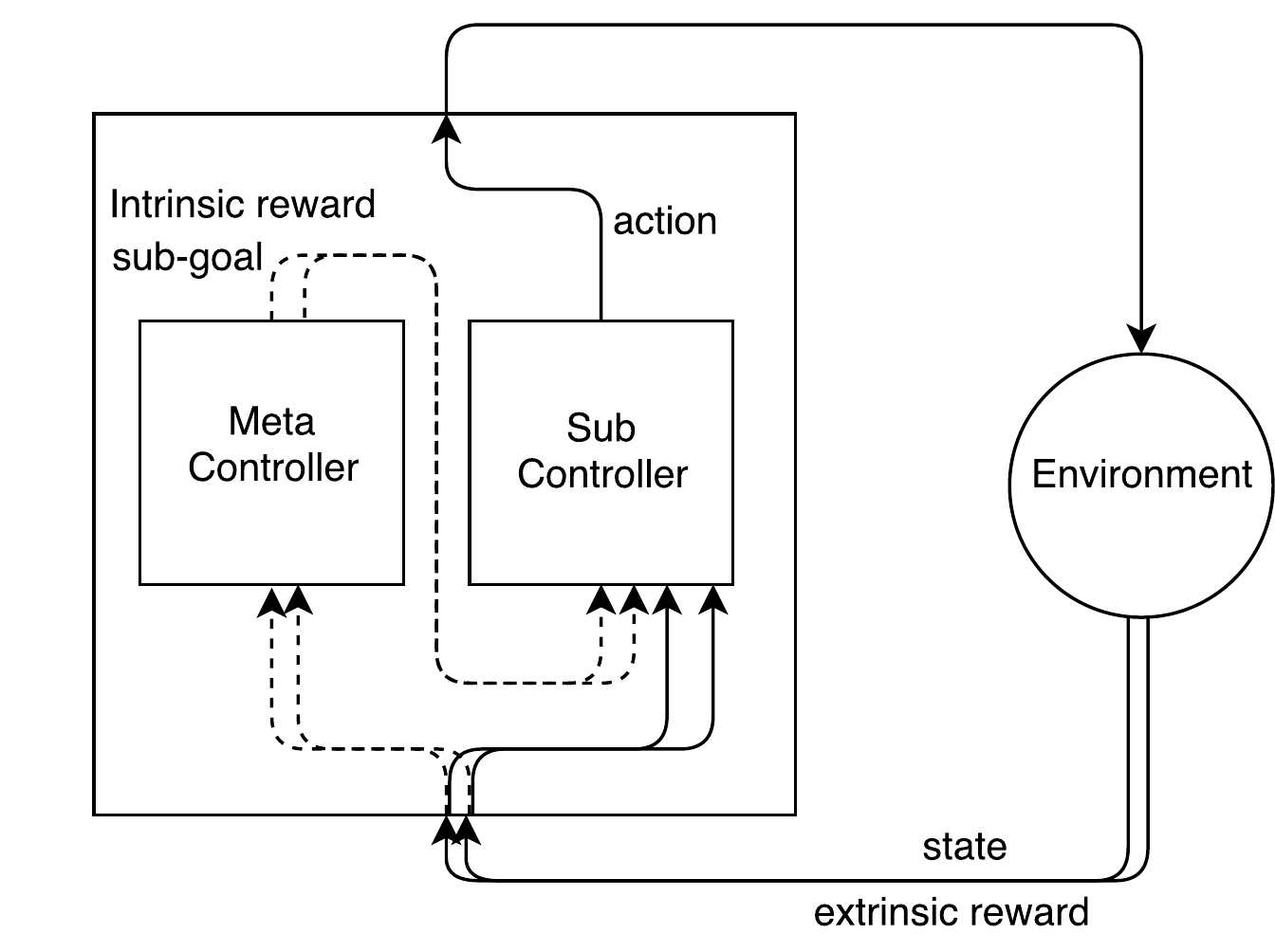}
    \caption{HRL action-perception loop}
    \label{fig:HRL}
	\end{subfigure}%
  \begin{subfigure}{0.2\textwidth}
	\centering
	\includegraphics[trim={5cm 0.0cm 5cm 1.0cm},clip,width = \textwidth ]{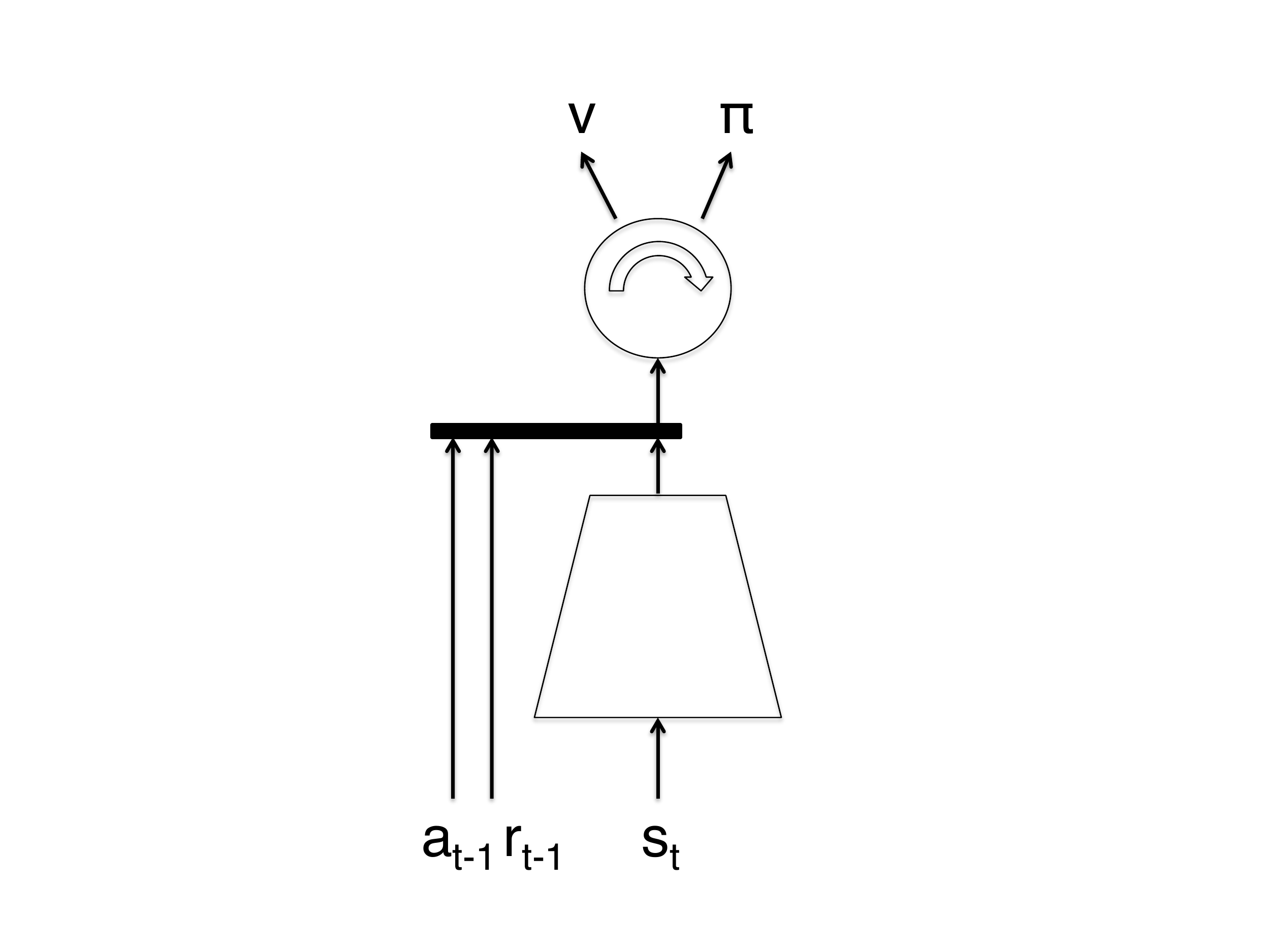}
    \caption{A3C baseline}
\end{subfigure}%
\begin{subfigure}{0.2\textwidth}
	\centering
	\includegraphics[trim={5cm 0.0cm 5cm 1.0cm},clip,width = \textwidth]		{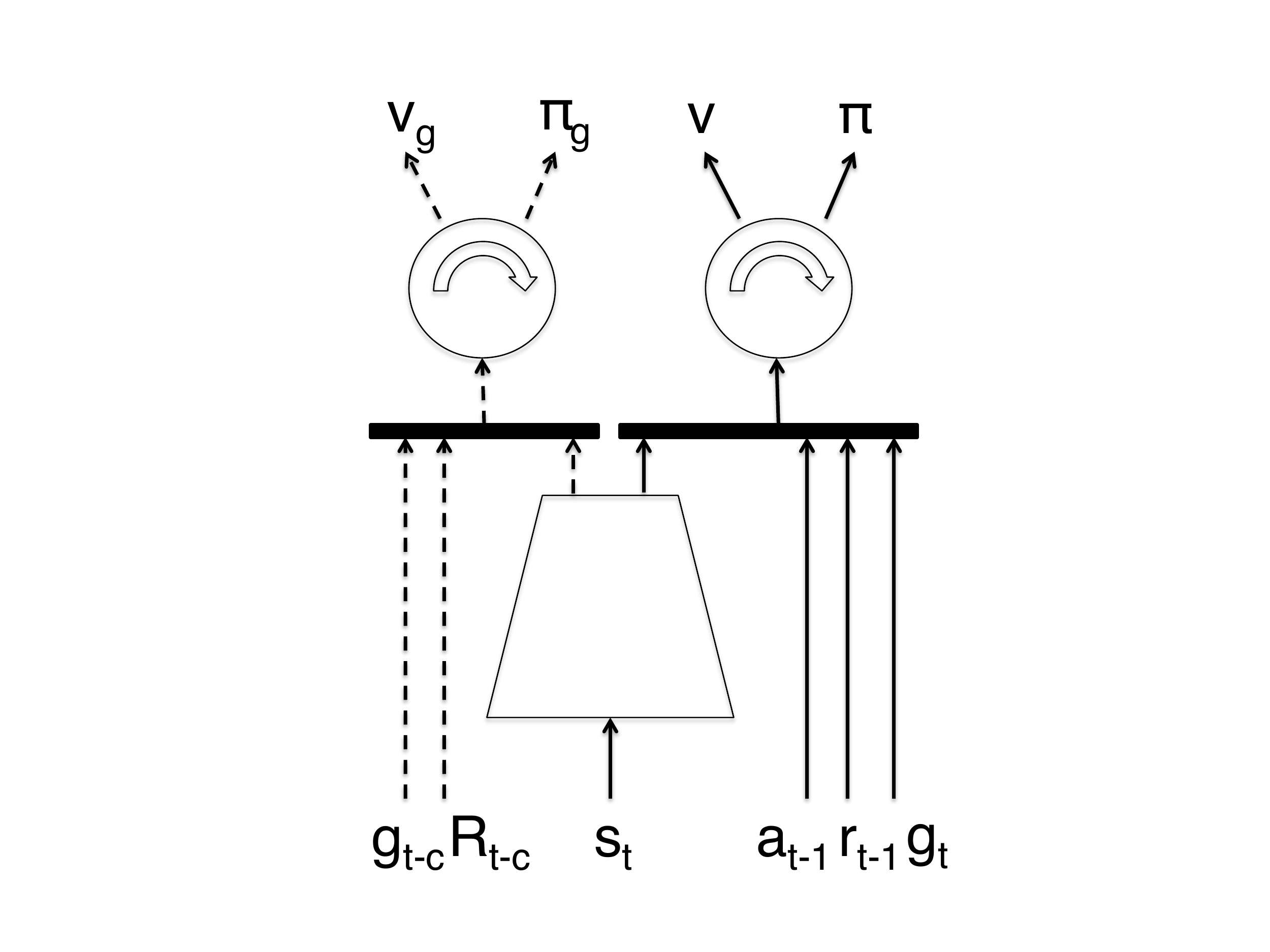}
    \caption{Our model}
	\end{subfigure}%

  \caption{(a) Depiction of the action-perception loop in our HRL system. (b) Diagram of the architecture of the A3C model used as a baseline. LSTM components are depicted with circles while the encoding layers are depicted with the trapezium (see Section~\ref{sec:architecture}). The bold horizontal line depicts a concatenation of all the incoming vectors. (c) Diagram of our proposed model. The most notable difference is an additional LSTM component which parameterises the meta-controller's policy and value function. The dotted line represents the slower time scale at which the meta-controller operates.}
  \label{fig:model}
\end{figure}

We consider the standard reinforcement learning setting where an agent interacts with an environment by observing the state of the environment $s_t$ and taking an action $a_t$ at every discrete time step $t$. The environment provides an extrinsic reward to the agent $r_t^{ext}$ and then transitions to the next state $s_{t+1}$. The goal of the agent is to maximise the accumulated sum of extrinsic rewards over the finite horizon length of an episode.

Specifically, we consider a hierarchical agent with two components: a meta-controller and a sub-controller. The sub-controller is responsible for choosing the agent's actions and directly interacts with the environment. The meta-controller operates on a longer time scale of $c$ time steps and influences the behaviour of the sub-controller through a \emph{subgoal argument}, $g_t$. This influence is imposed by giving $g_t$ as an input to the sub-controller in addition to $s_t$. Importantly, the meta-controller also gives intrinsic reward to the sub-controller, $r_t^{int}$, for successfully completing the subgoal $g_t$ and thus, by learning to associate $g_t$ and $r_t^{int}$, the sub-controller's behaviour is biased to complete the subgoals. At the same time, the meta-controller learns to select sequences of $g_t$ such that the sub-controller trajectory maximises accumulated extrinsic reward.

\subsection{Measure of control for intrinsic reward}
\label{sec:measure_control}
Here we detail two variants of our algorithm, corresponding to two ways to deliver the subgoal $g_t$: (a) the pixel-control agent and (b) the feature-control agent. Both agents have the same architecture, which will be discussed in Section \ref{sec:architecture}. The main difference between the two is the calculation of intrinsic reward, which is crucial for manipulation of the behaviour of the sub-controller.

\paragraph{Pixel control}

Following Jaderberg et al., we study the most basic form of controlling ability in the visual domain, which is the ability to control a given subset of pixels in the visual input~\cite{jaderberg2016reinforcement}. We divide the pre-processed 84x84 input image into 36 pixel patches of size 14x14. We define the intrinsic reward as the squared difference between two consecutive frames of pixels in the patch, normalized by the squared difference of the whole image. The sub-controller is thus encouraged to maximise the change in values of pixels in the given ($k^{th}$) patch relative to the entire screen. This can be written formally as
\begin{align}
 r^{int}(k) = \eta \frac{ ||\pmb{h}_k\odot (\pmb{s}_{t} - \pmb{s}_{t-1}) ||^2 }{ ||\pmb{s}_{t} - \pmb{s}_{t-1}||^2},
\end{align}
where $\pmb{h}_k$ is an 84x84 binary filter matrix with entries all equal to 0 apart from the $k^{th}$ pixel patch, which has entries all equal to 1.  By applying this filter with element-wise multiplication $\odot$, only the changes in the relevant part of the screen are taken into account. $\eta$ is a scaling factor which controls the magnitude of the intrinsic reward per time step.\footnote{We choose $\eta = 0.05$ which gives a reasonable value of accumulated intrinsic reward over an episode at the start of the training. We leave the tuning of this parameter for future work. While we believe tuning this parameter is important, the more important parameter is the relative weight between extrinsic and intrinsic reward, see eq.~\ref{eq:shaped_reward}.} 

\paragraph{Feature control}

Jaderberg et al. introduced a notion of feature control which is defined as the ability to control the activations of specific neurons. Similarly, Bengio et al. introduced the notion of feature selectivity that measures how much a feature can be controlled, independently from other features~\cite{bengio2017independently}. We define intrinsic reward as Bengio et al.'s feature selectivity measure on the second convolutional layer of our network. To measure the selectivity of a feature, we take the difference between the mean activation of a selected feature map at consecutive time steps and normalize with all feature maps. This can be written as
\begin{align}
r^{int}(k) = \eta \frac{||f_k(\pmb{s}_t) - f_k(\pmb{s}_{t-1})||}{\sum_{k'}||f_{k'}(\pmb{s}_t) - f_{k'}(\pmb{s}_{t-1})||},
\end{align}
where $f_k(.)$ denotes the mean over activation values in the $k^{th}$ feature map and $\sum_{k'}$ denotes summation over all feature maps. 

In contrast with the pixel-control agent, allowing the meta-controller to select a convolutional feature endows the agent with more flexible and abstract control of its environment. The instruction from the meta-controller is more abstract since a feature map can represent a complex function of the sensory inputs, and it is more flexible because the feature maps can be shaped during learning to encode aspects of the environment that are useful to control for the completion of the main task.


\paragraph{Shaped reward}
In addition to intrinsic reward, we also give extrinsic reward to the sub-controller, enabling it to learn fine-grained behaviour. We adjust the ratio between intrinsic and extrinsic reward with a parameter $\beta$, which results in the shaped reward, 
\begin{align}
\label{eq:shaped_reward}
r_t = \beta  r_t^{ext} + (1 - \beta) r_t^{int}.
\end{align}
The sub-controller strictly follows the order of the meta-controller if $\beta = 0$. On the other hand, if $\beta = 1$, the meta-controller has little direct influence on the sub-controller. In this case, the sub-controller still receives the subgoal argument as an input but does not receive rewards for attaining the subgoal.

\subsection{Architectural and optimisation details}
\label{sec:architecture}

Our baseline model is a variant of the asynchronous advantage actor-critic algorithm (A3C)~\cite{mnih2016asynchronous}. We adapted OpenAI's A3C implementation\footnote{Our implementation is an adaptation of an open-source implementation of A3C, namely, ``Universe-Starter-Agent''. (https://github.com/openai/universe-starter-agent) which is written with Tensorflow~\cite{tensorflow2015-whitepaper}}
\footnote{The source code of our implementation will be made publicly available.} to follow the architecture specified by Wang et al.~\cite{wang2016learning}. The model consists of two parts: an encoding module and a Long-Short Term Memory layer (LSTM)~\cite{hochreiter1997long}. The encoding module consists of two convolution layers and a fully connected layer. The first convolution has 16 8x8 filters with a stride length of 4 and the second layer has 32 4x4 filters with a stride length of 2. The second convolution is followed by a fully connected layer with 256 units. The output of the fully connected layer is then concatenated with the previous action and the previous reward, and then fed into an LSTM layer. The LSTM has 256 cells whose output linearly projects into the policy and value networks.

Our hierarchical model extends the baseline model as follows (Figure~\ref{fig:model}). An additional LSTM layer is added to parameterise the meta-controller's value and policy function. The input to the meta-controller's LSTM includes the previous subgoal argument and extrinsic rewards accumulated from the previous meta-step. The sub-controller's LSTM also has an additional input consisting of the current subgoal argument. The meta-controller operates every $c=100$ time steps.

The subgoal argument is a one-hot vector, which specifies the index of the subgoal selected by the meta-controller. For the pixel-control agent, there are 37 subgoals corresponding to 36 patches of pixel plus a no-op which gives no intrinsic reward. The feature-control agent has 32 discrete subgoals, each corresponding to a feature map in the second convolutional layer. 

The value and policy networks are optimised using the A3C loss function with 8 asynchronous agents, where the advantage is estimated with the generalized advantage estimator~\cite{schulman2015high}, using $\gamma = 0.99$ and $\lambda = 1.0 $. We use the ADAM optimizer~\cite{kingma2014adam} with a learning rate of $0.0001$ for all of our experiments. Our experiments use a backpropagation through time (BPTT) trajectory length of either 20 or 100 time steps for sub-controller and the baseline agent, and 20 meta-steps (2000 time steps) for the meta-controller. Finally, the gradients are scaled down when their $L_2$-norms exceed 40.

\section{Experimental results and discussion}

The objectives of our experiments are:
\begin{enumerate}
\item To verify that the influence from the meta-controller is beneficial in sparse reward environments.
\item To evaluate the performance of our agent in several environments in comparison to current state-of-the-art HRL systems.
\item To study the behaviour of the agents under the influence of pixel-control and feature-control intrinsic motivation.
\end{enumerate}

\subsection{The environments}

We evaluated the model on Atari games in the OpenAI gym environment~\cite{brockman2016openai}, a toolkit for comparing reinforcement learning algorithms that wraps the Arcade Learning Environment (ALE)~\cite{bellemare2013arcade} with a number of modifications. In our experiments, we make comparisons with the Feudal Network (FuN)~\citep{vezhnevets2017feudal} and Option-Critic ~\cite{bacon2016option} architectures, both evaluated on the ALE. \footnote{However, it is important to note that they are not directly comparable. OpenAI's gym adds stochasticity through the use of random frame-skips, while ALE is a deterministic environment. The standard evaluation protocol in ALE is to add a random number of no-op actions at the start of the episode to achieve some stochasticity.} The environment provides the state as 210x160x3 RGB pixels. We pre-process the state by reshaping it into an 84x84x3 matrix, retaining the RGB channels. We also clip the extrinsic reward to the range of [-1, 1]. We used v0 setting for all the games, e.g., MontezumaRevenge-v0 for Montezuma's Revenge.

\subsection{Experiment 1: Influence of the meta-controller on performance}
\begin{figure*}[t]
	\begin{subfigure}[t]{\textwidth}
    	\centering
    	\includegraphics[trim={2.0cm 0.8cm 0.8cm 0.8cm},clip, width=0.8\linewidth]{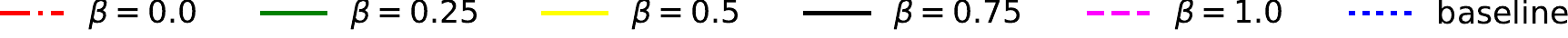}
    \end{subfigure}
    \begin{subfigure}[t]{0.25\textwidth}
        \includegraphics[trim={0.2cm 0.2cm 0.2cm 0.2cm},clip,width=0.9\linewidth]{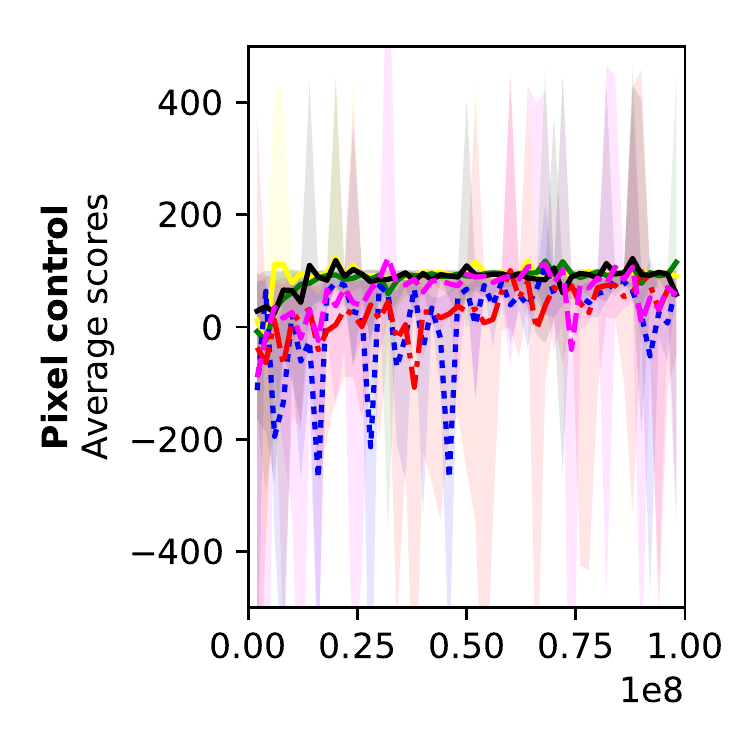}
    \end{subfigure}%
    \begin{subfigure}[t]{0.25\textwidth}
        \includegraphics[trim={0.2cm 0.2cm 0.2cm 0.2cm},clip,width=0.9\linewidth]{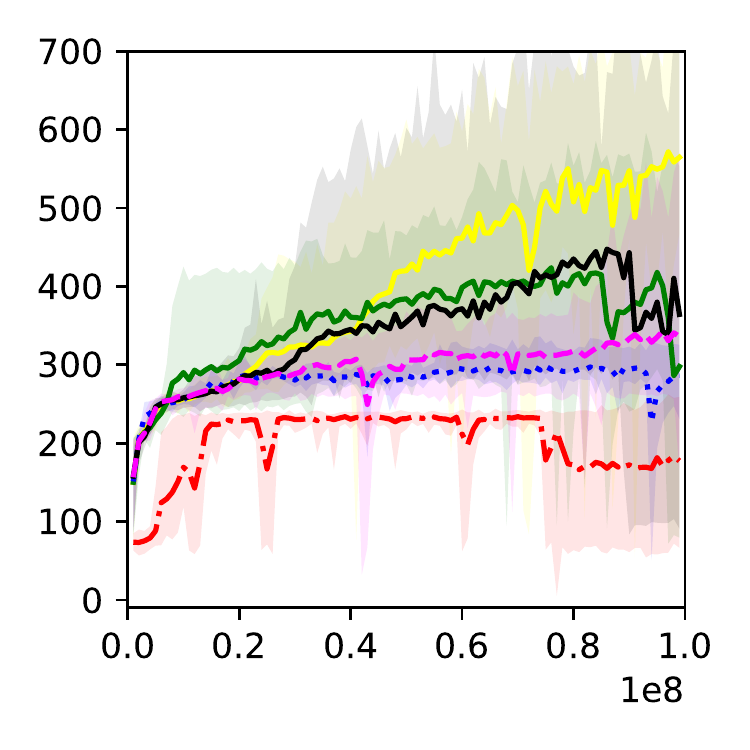}
    \end{subfigure}%
    \begin{subfigure}[t]{0.25\textwidth}
        \includegraphics[trim={0.2cm 0.2cm 0.2cm 0.2cm},clip,width=0.9\linewidth]{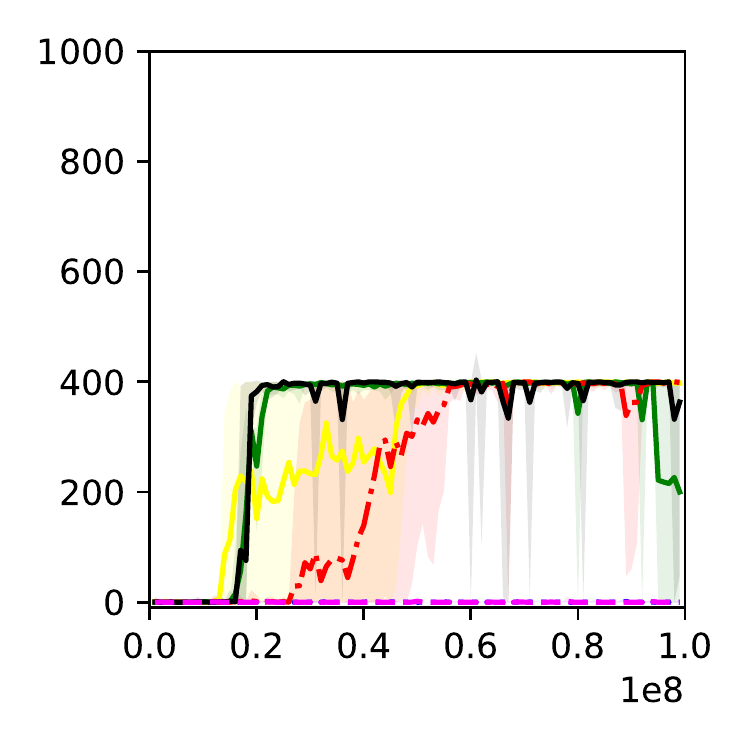}
    \end{subfigure}%
    \begin{subfigure}[t]{0.25\textwidth}        
        \includegraphics[trim={0.2cm 0.2cm 0.2cm 0.2cm},clip,width=0.9\linewidth]{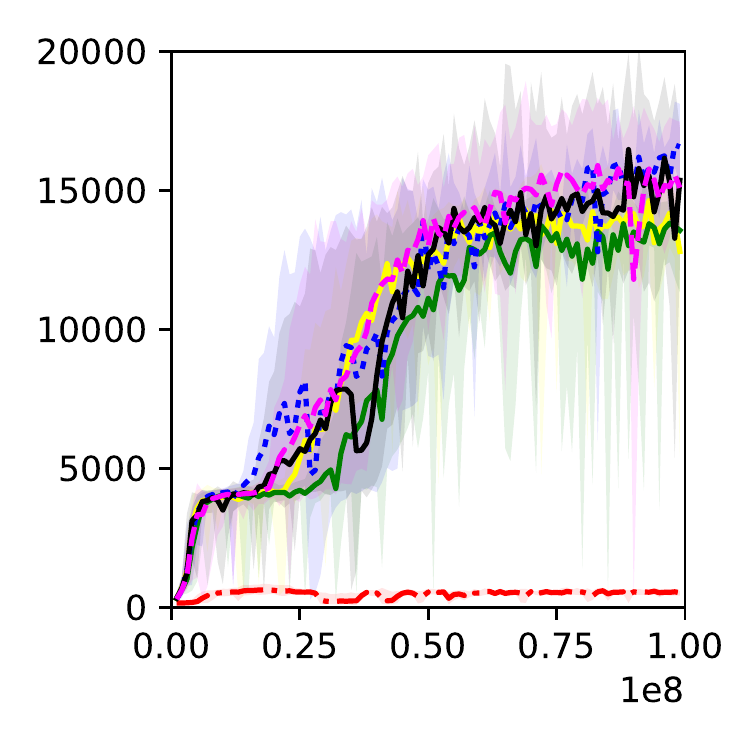}
    \end{subfigure}
    \begin{subfigure}[t]{0.25\textwidth}
        \includegraphics[trim={0.2cm 0.2cm 0.2cm 0.2cm},clip,width=0.9\linewidth]{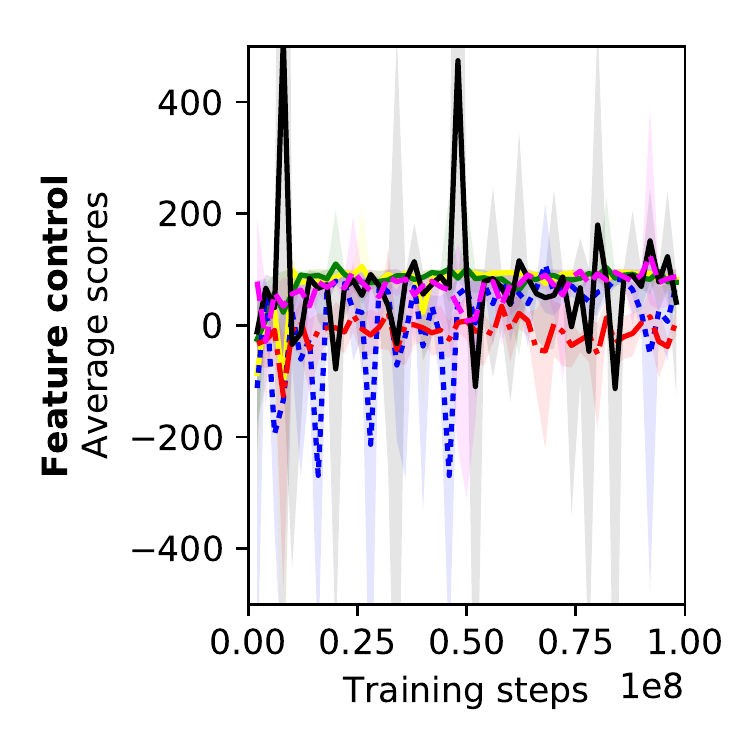}
        \caption{Private Eye}
    \end{subfigure}%
    \begin{subfigure}[t]{0.25\textwidth}
        \includegraphics[trim={0.2cm 0.2cm 0.2cm 0.2cm},clip,width=0.9\linewidth]{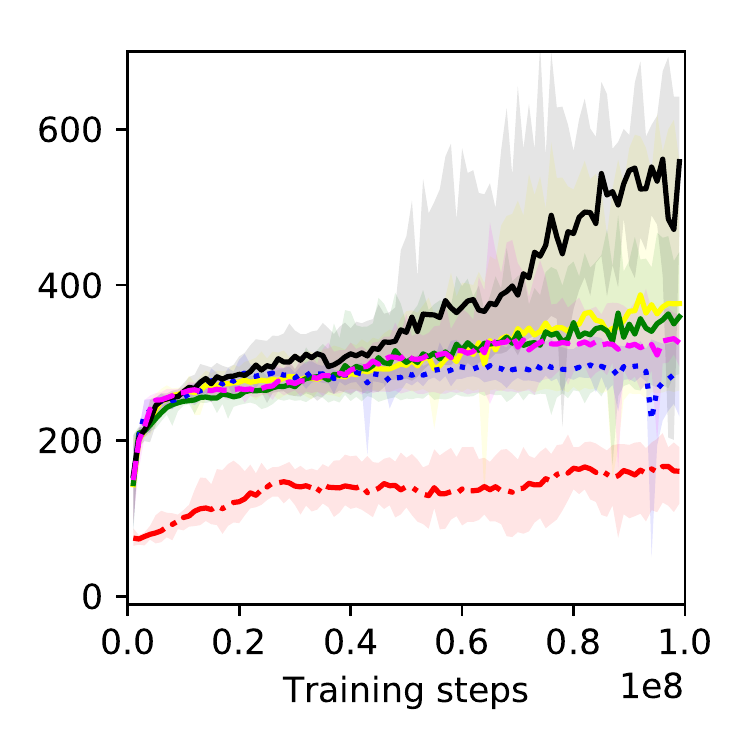}
        \caption{Frostbite}
    \end{subfigure}%
    \begin{subfigure}[t]{0.25\textwidth}
        \includegraphics[trim={0.2cm 0.2cm 0.2cm 0.2cm},clip,width=0.9\linewidth]{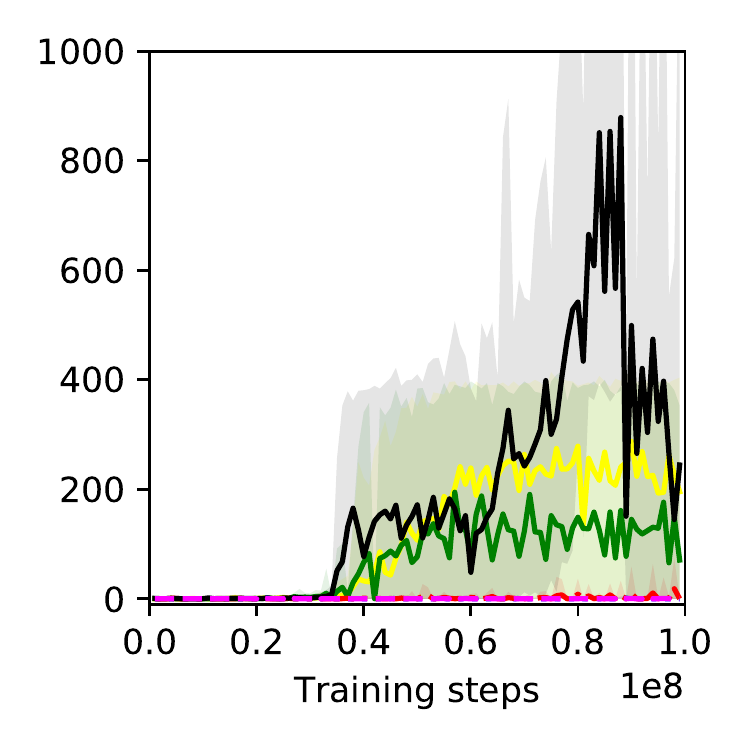}
        \caption{Montezuma's Revenge}
    \end{subfigure}%
    \begin{subfigure}[t]{0.25\textwidth}        
        \includegraphics[trim={0.2cm 0.2cm 0.2cm 0.2cm},clip,width=0.9\linewidth]{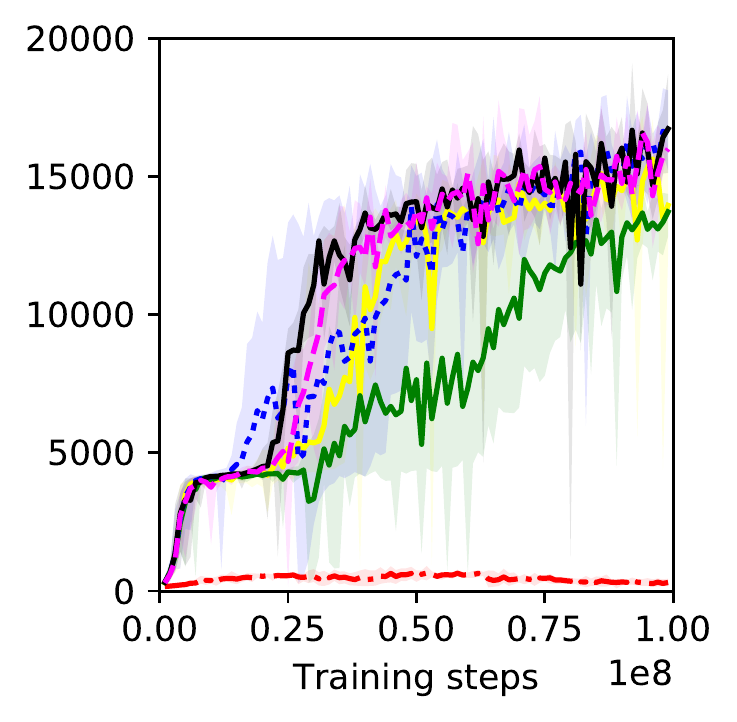}
        \caption{Q*bert}
    \end{subfigure}
    \caption{Performance of the pixel-control agent (\textbf{top row}) and the feature-control agent (\textbf{bottom row}) using different values of $\beta$. The vertical axis shows the average score per episode, evaluated after every 100 episodes. Each learning curve is an average over four runs with the shaded areas showing the difference between the maximum and minimum scores of the four runs.}
    \label{fig:exp_beta}
\end{figure*}


To evaluate the effectiveness of the meta-controller, we ran our agent with different relative weights $\beta \in \{0.00, 0.25, 0.50, 0.75, 1.00\}$ between extrinsic and intrinsic reward, and compared the performance with the baseline agent. We used BPTT = 20 for the sub-controller. First, we found that with $\beta = 1.00$ (no intrinsic reward) the agent's performance was similar to the baseline. This result demonstrates that any significant gain or decline in performance using other values of $\beta$ can be attributed to the intrinsic reward. 

As shown in Figure~\ref{fig:exp_beta}, the feature-control agent with $\beta = 0.75$ outperforms other agents in Montezuma's Revenge and Frostbite and is competitive with other agents in Q*bert and Private Eye. This result suggests that introducing a certain proportion of intrinsic reward in the sub-controller has a positive effect in sparse reward environments (such as Montezuma's Revenge) without degrading the performance on dense reward environments (such as Q*bert).

Our agents with $\beta = 0$ (no extrinsic reward) perform very poorly as expected. Since the sub-controller can only follow a limited number of subgoals from the meta-controller, its behaviours are also limited in this case. Giving extrinsic reward to the sub-controller is a way to allow fine-grained behaviours that are important for maximising extrinsic reward. Interestingly, agents with $\beta = 0.25$ also perform worse than baseline in Q*bert. This result shows that too much influence from the meta-controller can have negative effect in dense reward environments. Interestingly, the best value of $\beta$ is consistent across all four games.

We observe that the pixel-control agent learns more quickly than the feature-control agent. However, the feature-control agent generally achieves better scores after 100 million frames of training. This is likely due to the fact that the feature-control agent needs time to learn useful features before the influence of the meta-controller becomes meaningful. Once the features have been learned, the subgoals obtained are of higher quality than the hard-coded ones in the pixel-control agent.  

\subsection{Experiment 2: Performance instability and the length of the BPTT roll-out}

\begin{figure}
\centering
	\begin{subfigure}[t]{0.15\textwidth}
        \hspace{1em}
    \end{subfigure}%
	\begin{subfigure}[t]{0.3\textwidth}
    	\centering
        \includegraphics[width=\linewidth]{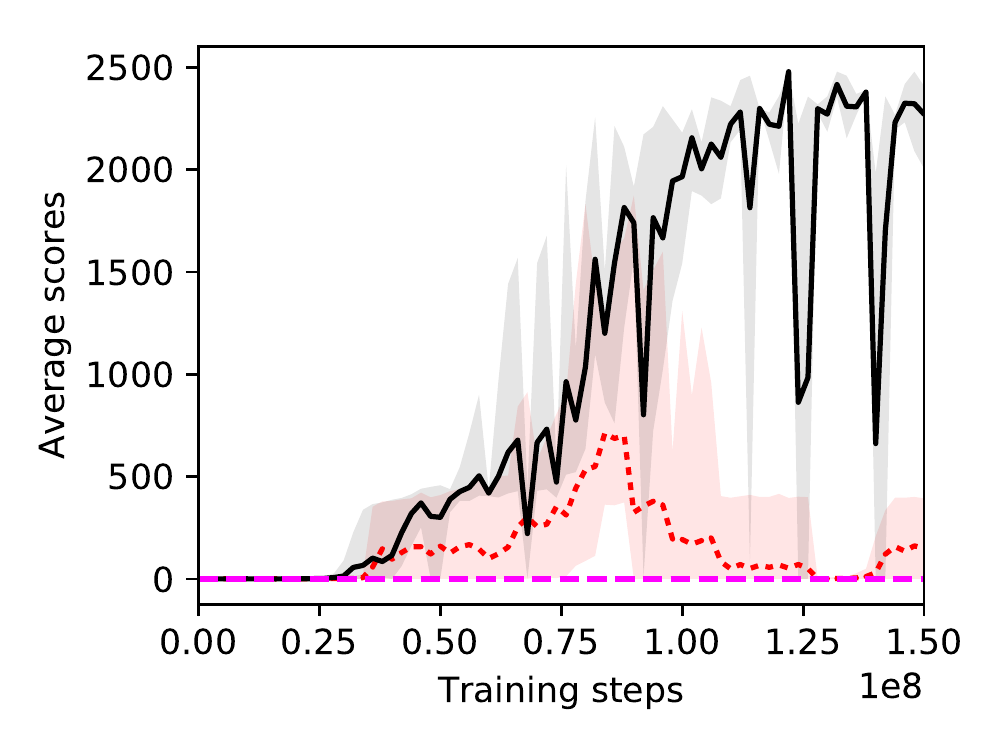}
        \caption{Montezuma's Revenge}
    \end{subfigure}%
	\begin{subfigure}[t]{0.3\textwidth}
    	\centering
        \includegraphics[width=\linewidth]{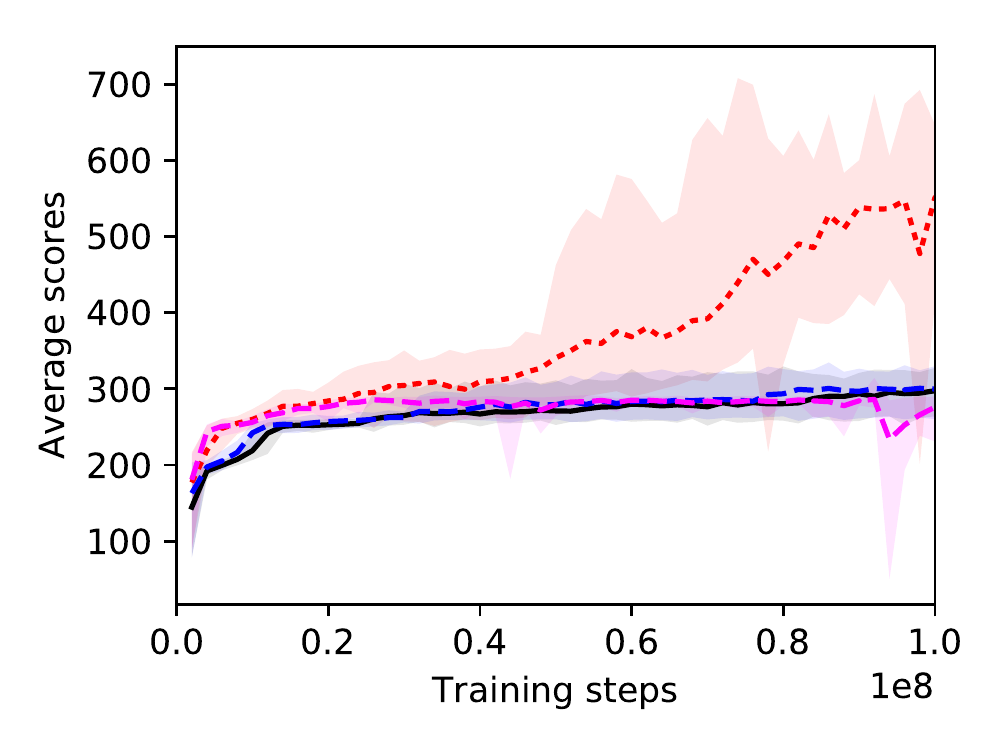}
        \caption{Frostbite}
    \end{subfigure}%
    \begin{subfigure}[t]{0.15\textwidth}
    	\centering
        \includegraphics[width=\linewidth, trim={1cm, 2cm, 2cm, 0cm}, clip]{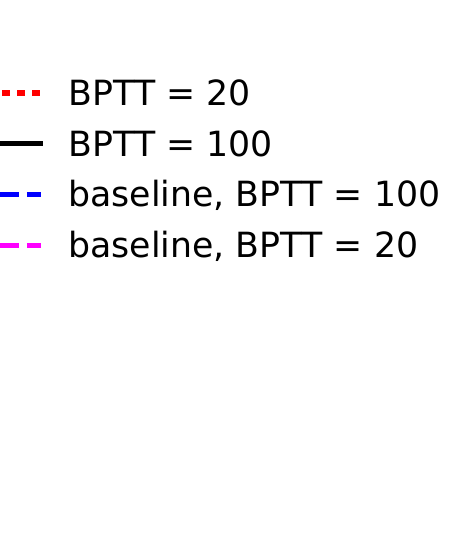}
    \end{subfigure}
\caption{Performance of the feature-control agent with $\beta = 0.75$ on Montezuma's Revenge (\textbf{left}) and Frostbite (\textbf{right}) with BPTT = 20 (dotted red line) and BPTT=100 (solid black line). We see a significant performance gain in Montezuma's Revenge when BPTT is increased to 100, but in Frostbite, it results in a drop in performance to the level of the baseline.}
\label{fig:montezuma_fun}
\end{figure}

In our initial experiments, we observed instability in the training curve of the feature-control agent, which comes in the form of catastrophic drops in performance. To alleviate this problem, we tried increasing the BPTT roll-out from 20 to 100 steps for the sub-controller. We reasoned that a longer unrolled sequence of BPTT could contribute to training stability in the following ways: (i) the updates are less frequent and give the agent more stable features, which are a crucial component in the calculation of the intrinsic reward, and (ii) it allows the gradient to be backpropagated further into the past, which potentially reduces bias in the update.


In Figure~\ref{fig:montezuma_fun} we see that in Montezuma's Revenge the agent attains a much higher score with BPTT = 100 than with BPTT = 20. In Frostbite, however, we observe the opposite effect. This could be because the gradient is already stable at BPTT = 20 and so increasing the BPTT length does not yield any positive effect; on the contrary, as a result of lowering the frequency of updates it can result in slower learning (see Figure \ref{fig:montezuma_fun}).

\subsection{Experiment 3: Comparison to state-of-the-art hierarchical RL systems}

In this experiment, we evaluated our feature-control agent on Ms. Pac-Man, Asterix, Zaxxon and Montezuma's Revenge. The aim was to show that the method is applicable to a broad range of games, and to compare our system to two state-of-the-art end-to-end HRL systems, namely the Option-Critic ~\cite{bacon2016option} and FuN~\cite{vezhnevets2017feudal} architectures. 

\begin{figure}
\centering
	\begin{subfigure}[t]{1.0\textwidth}
    	\centering
        \includegraphics[width=0.9\linewidth, trim={0.2cm, 0.2cm, 0.2cm, 0.2cm}, clip]{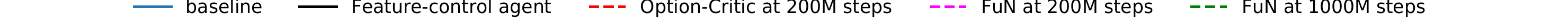}
    \end{subfigure}
	\begin{subfigure}[t]{0.25\textwidth}
        \includegraphics[width=0.9\linewidth, trim={0.2cm, 0.2cm, 0.2cm, 0.2cm}, clip]{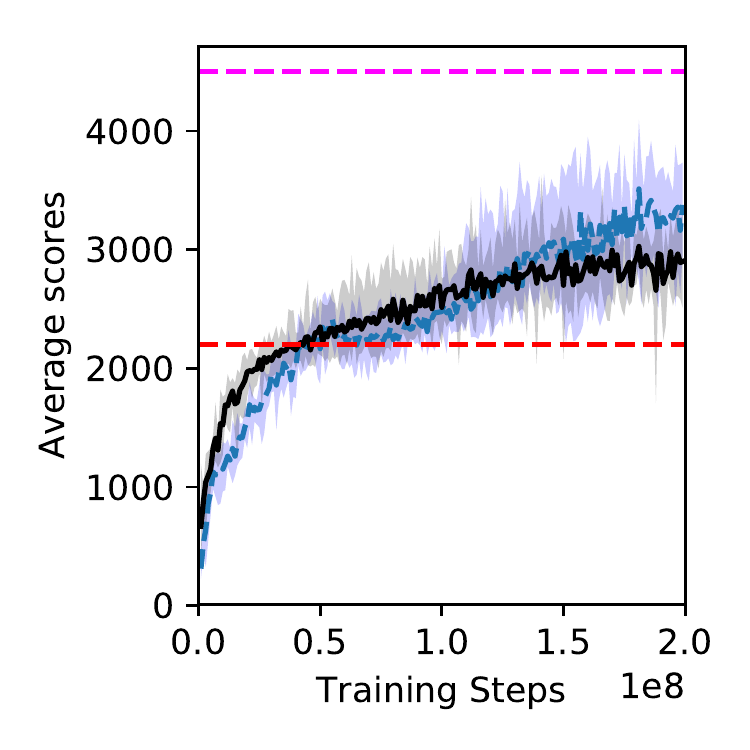}
        \caption{Ms. Pac-Man}
    \end{subfigure}%
	\begin{subfigure}[t]{0.25\textwidth}
        \includegraphics[width=0.9\linewidth, trim={0.2cm, 0.2cm, 0.2cm, 0.2cm}, clip]{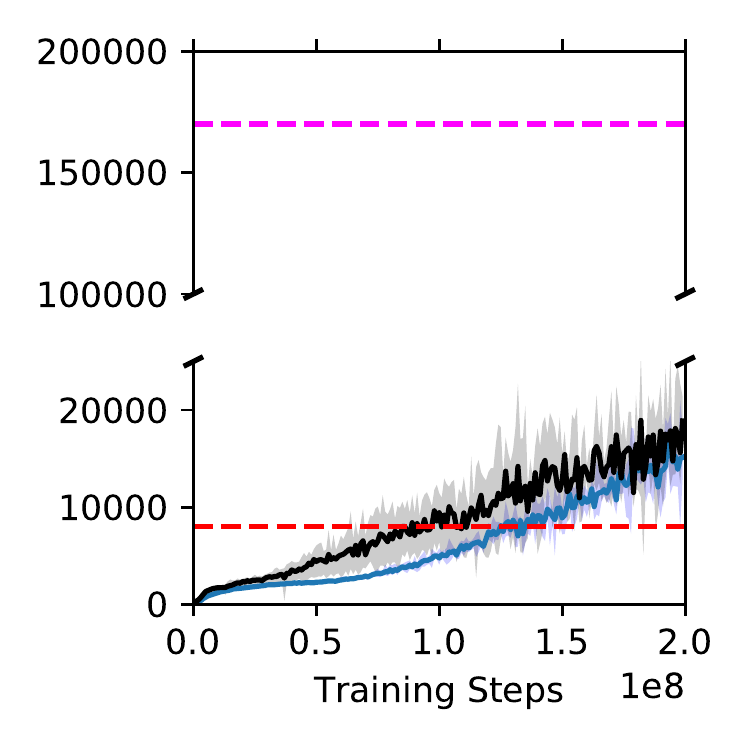}
        \caption{Asterix}
    \end{subfigure}%
    \begin{subfigure}[t]{0.25\textwidth}
        \includegraphics[width=0.9\linewidth, trim={0.2cm, 0.2cm, 0.2cm, 0.2cm}, clip]{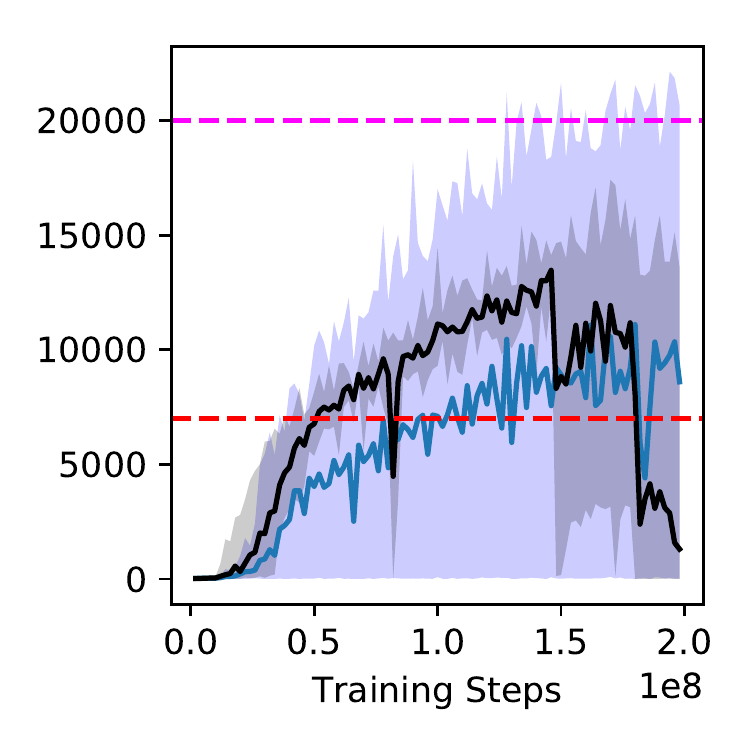}
        \caption{Zaxxon}
    \end{subfigure}%
    \begin{subfigure}[t]{0.25\textwidth}
        \includegraphics[width=0.9\linewidth, trim={0.2cm, 0.2cm, 0.2cm, 0.2cm}, clip]{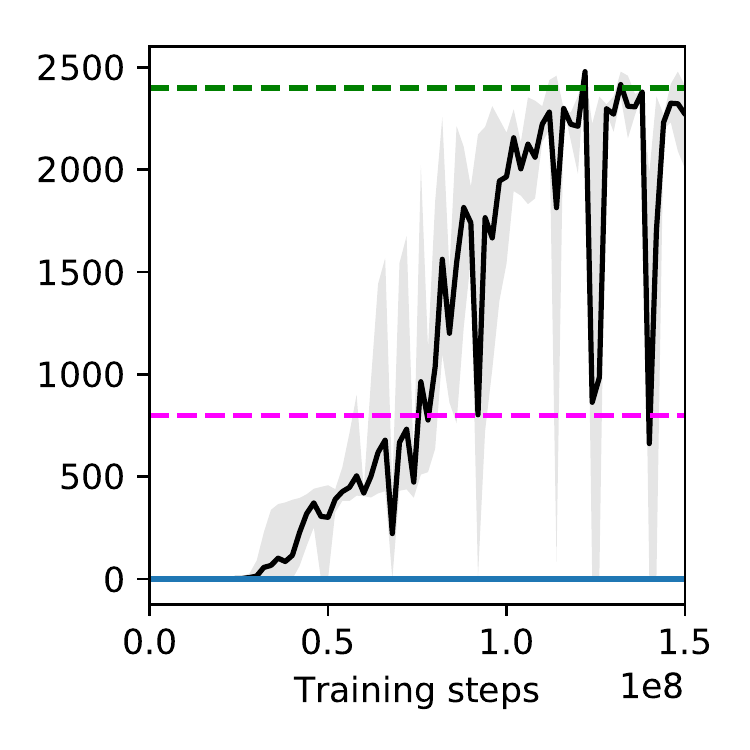}
        \caption{Montezuma's Revenge}
    \end{subfigure}
    
\caption{Comparisons of the feature-control agent with the FuN and Option-critic agents. $\beta$ is set to 0.75 for all games. We use BPTT = 20 for Ms. Pac-Man and Asterix, and BPTT = 100 for Zaxxon and Montezuma's Revenge. }
\label{fig:exp3}
\end{figure}

Our results are shown in Figure~\ref{fig:exp3} and we note the following: (i) On Ms. Pac-Man, Asterix and Zaxxon we achieve better maximum scores than the Option-Critic network but worse maximum scores than the FuN Network; (ii) on Montezuma's Revenge, our agent reaches approximately the same maximum score as the FuN network but it learns much more quickly, reaching this level of performance after fewer than a fifth of the number of observations. We anticipate being able to improve our agent's performance with a broader parameter search. For example, the discount parameter, $\gamma$, has been shown to have a significant impact on the performances of both A3C and FuN on different Atari games ~\cite{vezhnevets2017feudal}. We did not compare with other state-of-the-art results in Montezuma's Revenge, such as those obtained with the UNREAL agent~\cite{jaderberg2016reinforcement}, DQN-CTS~\cite{bellemare2016unifying} and DQN-PixelCNN~\cite{OstrovskiBOM17}, since these are not competing HRL methods and their advantageous features could easily be integrated into our agent.

\subsection{Experiment 4: Visualisation of the agent's behaviour under intrinsic motivation}

\begin{figure}
\centering
	\begin{subfigure}[t]{\textwidth}
    	\centering
        \includegraphics[width=\fsize\linewidth]{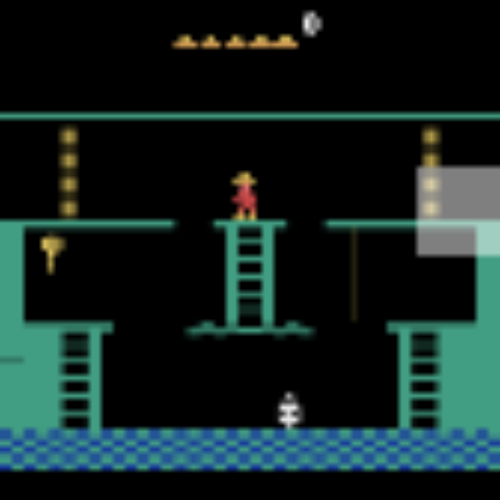}
        \includegraphics[width=\fsize\linewidth]{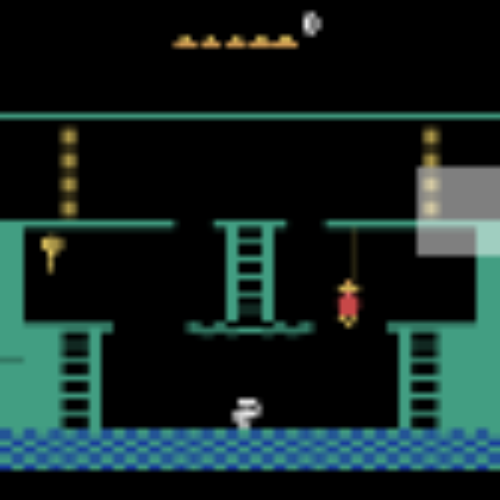}
        \includegraphics[width=\fsize\linewidth]{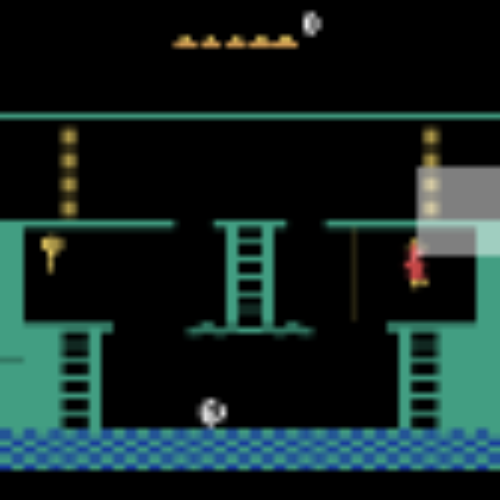}
        \includegraphics[width=\fsize\linewidth]{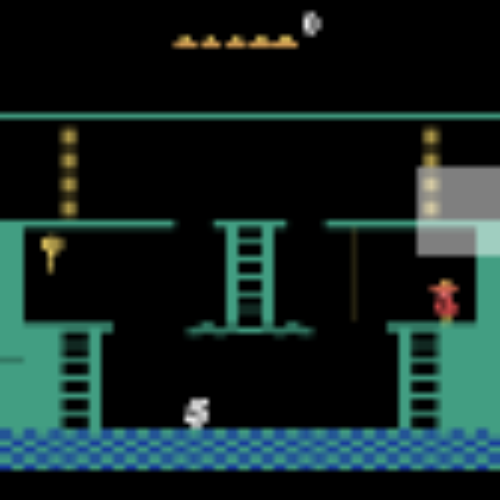}
        \includegraphics[width=\fsize\linewidth]{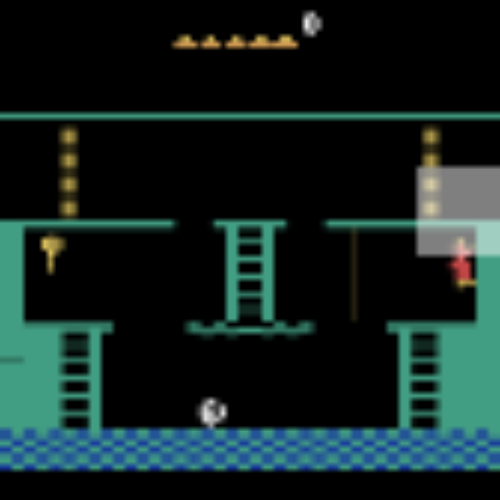}
        \caption{Pixel-control agent exploring the first room of Montezuma's Revenge}
        \label{fig:pixel_control}
    \end{subfigure}
    \\
    \begin{subfigure}[t]{\textwidth}
    	\centering
        \includegraphics[width=\fsize\linewidth]{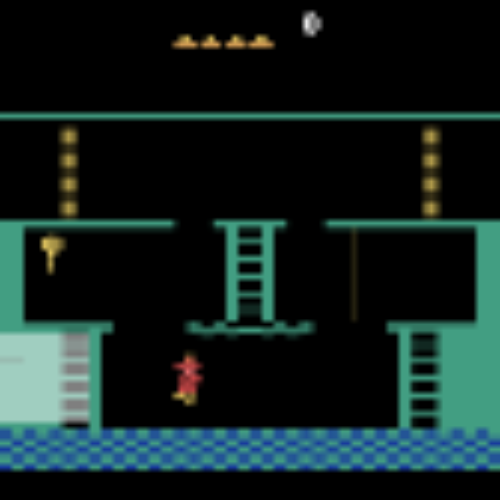}
        \includegraphics[width=\fsize\linewidth]{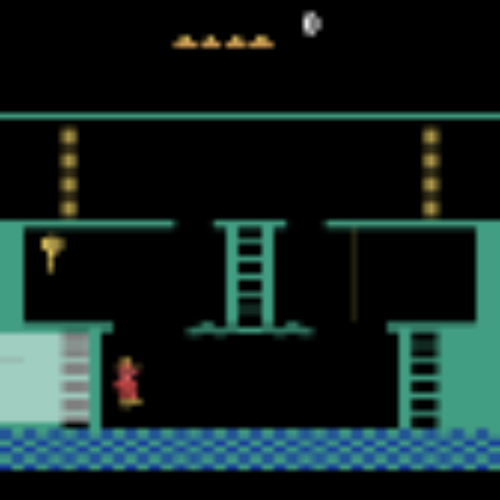}
        \includegraphics[width=\fsize\linewidth]{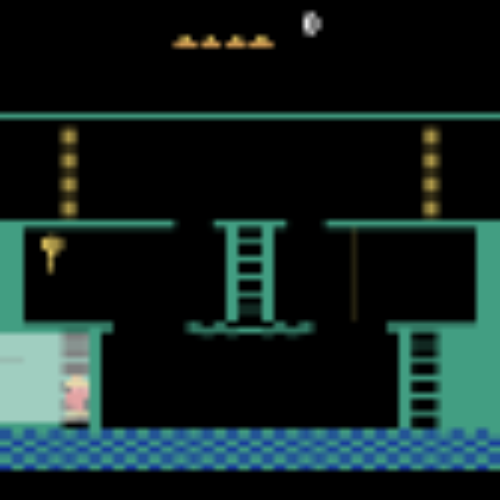}
        \includegraphics[width=\fsize\linewidth]{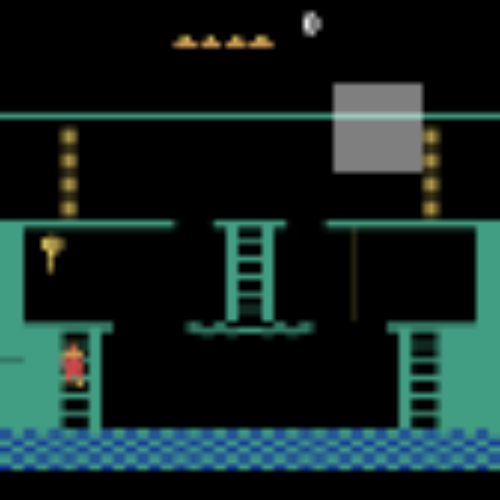}
        \includegraphics[width=\fsize\linewidth]{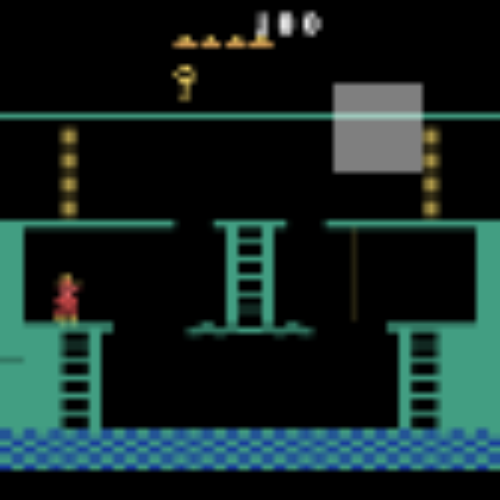}
        \caption{Pixel-control agent collecting the key in the first room of Montezuma's Revenge}
        \label{fig:pixel_control_key}
    \end{subfigure}
    \\
	\begin{subfigure}[t]{\textwidth}
    	\centering
        \includegraphics[width=\fsize\linewidth]{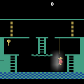}
        \includegraphics[width=\fsize\linewidth]{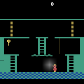}
        \includegraphics[width=\fsize\linewidth]{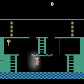}
        \includegraphics[width=\fsize\linewidth]{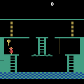}
        \includegraphics[width=\fsize\linewidth]{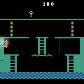}
        \caption{Feature-control agent collecting the key in the first room of Montezuma's Revenge}
        \label{fig:feature_control_r1}
    \end{subfigure}
    \\
    \begin{subfigure}[t]{\textwidth}
    	\centering
        \includegraphics[width=\fsize\linewidth]{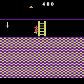}
        \includegraphics[width=\fsize\linewidth]{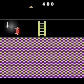}
        \includegraphics[width=\fsize\linewidth]{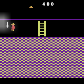}
        \includegraphics[width=\fsize\linewidth]{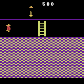}
        \includegraphics[width=\fsize\linewidth]{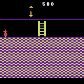}
        \caption{Feature-control agent collecting the sword in a room of Montezuma's Revenge}
        \label{fig:feature_control_r2}
    \end{subfigure}
\caption{Examples of screenshot sequences illustrating the intrinsic motivations of (a,b) the pixel-control agent and  (c,d) the feature-control agent. In (a) and (b), the white square represents the pixel patch chosen by the meta-controller and, in (c) and (d), the white cloud represents the upsampled activations of the chosen feature map.}
\label{fig:feature_control_montezuma}
\end{figure}

The influence of the intrinsic motivation provided by the meta-controller on the agent's behaviour can be most easily visualised with the pixel-control agent. In Figure \ref{fig:feature_control_montezuma}a, a sequence of screenshots from Montezuma's Revenge is shown where we see the sub-controller moving the character to the patch selected by the meta-controller and causing it to jump around in the patch in order to generate intrinsic reward. Figure \ref{fig:feature_control_montezuma}b shows another sequence where the meta-controller selects a patch over the ladder that must be climbed to collect the key. The character moves to the patch, but when the meta-controller then changes the location of the patch, the sub-controller ignores it and instead proceeds to collect the key, which results in extrinsic reward. This example highlights the importance of motivating the sub-controller with extrinsic as well as intrinsic reward, allowing the agent to be flexible and not completely at the mercy of the meta-controller.


Interpreting the intrinsic motivation of the feature-control agent is much more difficult, since it involves understanding what is encoded in the selected convolutional feature map. In an attempt to visualise this, we upsampled the selected feature map and overlaid it with the raw state input\footnote{This approach of visualisation of attention features is used by Xu et al. (2015)~\cite{xu2015show}.}. The feature-control agent has to learn strategies to maximally change the activations of this feature map to gain intrinsic reward.

We present two scenarios with the feature-control agent in Figure \ref{fig:feature_control_montezuma}c and d, which indicate how different types of features can evolve to form useful options for the agent. Figure \ref{fig:feature_control_r1} shows the agent collecting the key in the first room of Montezuma's Revenge. In this scenario, the feature map is activated in front of the agent on the path towards the key. This implicitly encourages the agent to move towards the key, as it attempts to maximally alter the activations of the feature map. Figure \ref{fig:feature_control_r2} shows the agent collecting the sword in another room. In this scenario, the feature map is only activated when the agent completes the apparent sub-task (collecting the sword), as opposed to the first scenario where the entire path to completing the sub-task (collecting the key) is highlighted.

\section{Conclusion}

In this paper, we presented an approach to tackling the reward sparsity problem in the form of a two-module deep hierarchical agent. In Montezuma's Revenge, an Atari game with particularly sparse rewards, our agent learns several times faster than the current state-of-the-art HRL agents, reaching a similar final level of performance. We also show that our subgoal designs are generically applicable across visual tasks by evaluating the agent on several different games. Our agent almost always performs better than the baseline agent, which suggests that the ability to control aspects of the environment can be a generically useful subgoal.

We argue that part of the performance gain comes from the ability to perform temporally-abstracted exploration. By visualising the trajectories of the pixel-control agent, we observe that it successfully learns to move towards the patch selected by the meta-controller in order to maximise its intrinsic reward; the acquisition of this skill allows the meta-controller to motivate the agent to explore its environment in a broad and temporally extended manner. In the feature-control agent, the options are learned via the shaping of the convolutional features and, while the features are harder to interpret than the pixel patches, there is evidence from our visualisations that they are activated by the completion of intuitive subgoals, such as collection of the sword in Montezuma's Revenge. 

An important result from our experiments is that the best performances are achieved when the sub-controller is motivated by a combination of intrinsic and extrinsic reward. By \emph{leaking} some extrinsic reward to the sub-controller, it frees us from the restriction that subgoals need to be complete or carefully designed, which can lead to brittle or sub-optimal solutions~\cite{dietterich2000hierarchical}. As long as the subgoals are useful for exploration, an agent equipped with such skills can learn faster, while still maintaining the ability to fine-tune its behaviour to maximise extrinsic reward.

In order to give our agent more flexibility, it would be interesting, in future work, to incorporate a termination condition to the options, which would allow the instruction from the meta-controller to be variable in length and thus more temporally precise. Additionally it would be interesting to quantify the extent to which our agents have learned to control their environment, perhaps by using the measure of empowerment~\cite{klyubin2005empowerment}.

\red{UPDATE 22/11/17: We later found that a flat A3C agent trained with shaped reward according to
equation 3 can perform as well as our feature-control agent in Montezuma Revenge. 
This result, in line with [13] and [27], supports the claim that additional auxiliary rewards or loss signals can be beneficial when dealing with
sparse reward environments even though the reward can possibly skew the definition of its task.}

\red{Importantly, this raises a question about the benefit of having the hierarchical elements proposed in this paper. It appears that decisions made by the meta-controller do not significantly contribute to the success of feature-control agent.
}

\subsubsection*{Acknowledgments}
We would like to thank Marc Deisenroth for providing us with Azure credits from the Microsoft Azure Sponsorship for Teaching and Research. We would also like to thank Kyriacos Nikiforou, Hugh Salimbeni and Kai Arulkumaran for fruitful discussions. N.D. is supported by the DPST scholarship from the Thai government.

\bibliographystyle{abbrv}
\bibliography{references.bib}

\begin{thebibliography}{10}

\bibitem{tensorflow2015-whitepaper}
M.~Abadi, A.~Agarwal, P.~Barham, E.~Brevdo, Z.~Chen, C.~Citro, G.~S. Corrado,
  A.~Davis, J.~Dean, M.~Devin, S.~Ghemawat, I.~Goodfellow, A.~Harp, G.~Irving,
  M.~Isard, Y.~Jia, R.~Jozefowicz, L.~Kaiser, M.~Kudlur, J.~Levenberg,
  D.~Man\'{e}, R.~Monga, S.~Moore, D.~Murray, C.~Olah, M.~Schuster, J.~Shlens,
  B.~Steiner, I.~Sutskever, K.~Talwar, P.~Tucker, V.~Vanhoucke, V.~Vasudevan,
  F.~Vi\'{e}gas, O.~Vinyals, P.~Warden, M.~Wattenberg, M.~Wicke, Y.~Yu, and
  X.~Zheng.
\newblock {TensorFlow}: Large-scale machine learning on heterogeneous systems,
  2015.
\newblock Software available from tensorflow.org.

\bibitem{bacon2016option}
P.-L. Bacon, J.~Harb, and D.~Precup.
\newblock The option-critic architecture.
\newblock In {\em AAAI Conference on Artificial Intelligence}, 2017.

\bibitem{barto2003recent}
A.~G. Barto and S.~Mahadevan.
\newblock Recent advances in hierarchical reinforcement learning.
\newblock {\em Discrete Event Dynamic Systems}, 13(4):341--379, 2003.

\bibitem{bellemare2016unifying}
M.~Bellemare, S.~Srinivasan, G.~Ostrovski, T.~Schaul, D.~Saxton, and R.~Munos.
\newblock Unifying count-based exploration and intrinsic motivation.
\newblock In {\em Advances in Neural Information Processing Systems}, pages
  1471--1479, 2016.

\bibitem{bellemare2013arcade}
M.~G. Bellemare, Y.~Naddaf, J.~Veness, and M.~Bowling.
\newblock The arcade learning environment: An evaluation platform for general
  agents.
\newblock {\em Journal of Artificial Intelligence Research}, 47:253--279, 2013.

\bibitem{bengio2017independently}
E.~{Bengio}, V.~{Thomas}, J.~{Pineau}, D.~{Precup}, and Y.~{Bengio}.
\newblock {Independently Controllable Features}.
\newblock {\em arXiv preprint arXiv:1703.07718}, 2017.

\bibitem{brockman2016openai}
G.~Brockman, V.~Cheung, L.~Pettersson, J.~Schneider, J.~Schulman, J.~Tang, and
  W.~Zaremba.
\newblock Openai gym.
\newblock {\em arXiv preprint arXiv:1606.01540}, 2016.

\bibitem{dayan1992feudal}
P.~Dayan and G.~E. Hinton.
\newblock Feudal reinforcement learning.
\newblock In {\em Proceedings of the 5th International Conference on Neural
  Information Processing Systems}, pages 271--278. Morgan Kaufmann Publishers
  Inc., 1992.

\bibitem{dietterich2000hierarchical}
T.~G. Dietterich.
\newblock Hierarchical reinforcement learning with the maxq value function
  decomposition.
\newblock {\em Journal of Artificial Intelligence Research}, 13:227--303, 2000.

\bibitem{gregor2016variational}
K.~{Gregor}, D.~{Jimenez Rezende}, and D.~{Wierstra}.
\newblock {Variational Intrinsic Control}.
\newblock {\em International Conference on Learning Representations Workshop},
  2017.

\bibitem{hochreiter1997long}
S.~Hochreiter and J.~Schmidhuber.
\newblock Long short-term memory.
\newblock {\em Neural computation}, 9(8):1735--1780, 1997.

\bibitem{houthooft2016vime}
R.~Houthooft, X.~Chen, Y.~Duan, J.~Schulman, F.~De~Turck, and P.~Abbeel.
\newblock Vime: Variational information maximizing exploration.
\newblock In {\em Advances in Neural Information Processing Systems}, pages
  1109--1117, 2016.

\bibitem{jaderberg2016reinforcement}
M.~Jaderberg, V.~Mnih, W.~M. Czarnecki, T.~Schaul, J.~Z. Leibo, D.~Silver, and
  K.~Kavukcuoglu.
\newblock Reinforcement learning with unsupervised auxiliary tasks.
\newblock In {\em International Conference on Learning Representations}, 2017.

\bibitem{kingma2014adam}
D.~Kingma and J.~Ba.
\newblock Adam: A method for stochastic optimization.
\newblock In {\em International Conference on Learning Representations}, 2015.

\bibitem{klyubin2005empowerment}
A.~S. Klyubin, D.~Polani, and C.~L. Nehaniv.
\newblock Empowerment: A universal agent-centric measure of control.
\newblock In {\em Evolutionary Computation, 2005. The 2005 IEEE Congress on},
  volume~1, pages 128--135. IEEE, 2005.

\bibitem{konidaris2009skill}
G.~Konidaris and A.~G. Barto.
\newblock Skill discovery in continuous reinforcement learning domains using
  skill chaining.
\newblock In {\em Advances in Neural Information Processing Systems}, pages
  1015--1023, 2009.

\bibitem{kulkarni2016hierarchical}
T.~D. Kulkarni, K.~Narasimhan, A.~Saeedi, and J.~Tenenbaum.
\newblock Hierarchical deep reinforcement learning: Integrating temporal
  abstraction and intrinsic motivation.
\newblock In {\em Advances in Neural Information Processing Systems}, pages
  3675--3683, 2016.

\bibitem{mcgovern2001automatic}
A.~McGovern and A.~G. Barto.
\newblock Automatic discovery of subgoals in reinforcement learning using
  diverse density.
\newblock In {\em Proceedings of the Eighteenth International Conference on
  Machine Learning}, pages 361--368. Morgan Kaufmann Publishers Inc., 2001.

\bibitem{menache2002q}
I.~Menache, S.~Mannor, and N.~Shimkin.
\newblock Q-cut—dynamic discovery of sub-goals in reinforcement learning.
\newblock In {\em European Conference on Machine Learning}, pages 295--306.
  Springer, 2002.

\bibitem{mnih2016asynchronous}
V.~Mnih, A.~P. Badia, M.~Mirza, A.~Graves, T.~Lillicrap, T.~Harley, D.~Silver,
  and K.~Kavukcuoglu.
\newblock Asynchronous methods for deep reinforcement learning.
\newblock In {\em International Conference on Machine Learning}, pages
  1928--1937, 2016.

\bibitem{mnih2015human}
V.~Mnih, K.~Kavukcuoglu, D.~Silver, A.~A. Rusu, J.~Veness, M.~G. Bellemare,
  A.~Graves, M.~Riedmiller, A.~K. Fidjeland, G.~Ostrovski, et~al.
\newblock Human-level control through deep reinforcement learning.
\newblock {\em Nature}, 518(7540):529--533, 2015.

\bibitem{mohamed2015variational}
S.~Mohamed and D.~J. Rezende.
\newblock Variational information maximisation for intrinsically motivated
  reinforcement learning.
\newblock In {\em Advances in Neural Information Processing Systems}, pages
  2125--2133, 2015.

\bibitem{OstrovskiBOM17}
G.~Ostrovski, M.~G. Bellemare, A.~van~den Oord, and R.~Munos.
\newblock Count-based exploration with neural density models.
\newblock In {\em International Conference on Machine Learning}, 2017.

\bibitem{parr1998reinforcement}
R.~Parr and S.~J. Russell.
\newblock Reinforcement learning with hierarchies of machines.
\newblock In {\em Advances in Neural Information Processing Systems}, pages
  1043--1049, 1998.

\bibitem{schmidhuber1991curious}
J.~Schmidhuber.
\newblock Curious model-building control systems.
\newblock In {\em Neural Networks, 1991. 1991 IEEE International Joint
  Conference on}, pages 1458--1463. IEEE, 1991.

\bibitem{schulman2015high}
J.~Schulman, P.~Moritz, S.~Levine, M.~Jordan, and P.~Abbeel.
\newblock High-dimensional continuous control using generalized advantage
  estimation.
\newblock In {\em International Conference on Learning Representations}, 2016.

\bibitem{csimcsek2004using}
{\"O}.~{\c{S}}im{\c{s}}ek and A.~G. Barto.
\newblock Using relative novelty to identify useful temporal abstractions in
  reinforcement learning.
\newblock In {\em Proceedings of the twenty-first international conference on
  Machine learning}, page~95. ACM, 2004.

\bibitem{Sutton:1998}
R.~S. Sutton and A.~G. Barto.
\newblock {\em Introduction to Reinforcement Learning}.
\newblock MIT Press, Cambridge, MA, USA, 1st edition, 1998.

\bibitem{sutton1999between}
R.~S. Sutton, D.~Precup, and S.~Singh.
\newblock Between mdps and semi-mdps: A framework for temporal abstraction in
  reinforcement learning.
\newblock {\em Artificial Intelligence}, 112(1-2):181--211, 1999.

\bibitem{vezhnevets2017feudal}
A.~S. Vezhnevets, S.~Osindero, T.~Schaul, N.~Heess, M.~Jaderberg, D.~Silver,
  and K.~Kavukcuoglu.
\newblock Feudal networks for hierarchical reinforcement learning.
\newblock In {\em International Conference on Machine Learning}, 2017.

\bibitem{wang2016learning}
J.~X. Wang, Z.~Kurth-Nelson, D.~Tirumala, H.~Soyer, J.~Z. Leibo, R.~Munos,
  C.~Blundell, D.~Kumaran, and M.~Botvinick.
\newblock Learning to reinforcement learn.
\newblock {\em arXiv preprint arXiv:1611.05763}, 2016.

\bibitem{xu2015show}
K.~Xu, J.~Ba, R.~Kiros, K.~Cho, A.~Courville, R.~Salakhudinov, R.~Zemel, and
  Y.~Bengio.
\newblock Show, attend and tell: Neural image caption generation with visual
  attention.
\newblock In {\em International Conference on Machine Learning}, pages
  2048--2057, 2015.

\end{thebibliography}

\end{document}